\documentclass[10pt,twocolumn,letterpaper]{article}

\usepackage{cvpr}              
\usepackage{multirow}
\usepackage{graphicx}
\usepackage[table]{xcolor}  
\usepackage{colortbl} 
\usepackage{booktabs}
\usepackage{caption}
\usepackage{array}
\usepackage{adjustbox}
\usepackage{siunitx}
\usepackage{tabularx}

\newcommand\blfootnote[1]{%
  \begingroup
  \renewcommand\thefootnote{}\footnote{#1}%
  \addtocounter{footnote}{-1}%
  \endgroup
}
\definecolor{cvprblue}{rgb}{0.21,0.49,0.74}
\usepackage[pagebackref,breaklinks,colorlinks,allcolors=cvprblue]{hyperref}

\title{SplitFlux: Learning to Decouple Content and Style from a Single Image}

\author{
    Yitong Yang$^{1}$\quad Yinglin Wang$^{1*}$\quad Changshuo Wang$^{2}$\quad Yongjun Zhang$^{3}$\quad Ziyang Chen$^{3}$ \quad Shuting He$^{1*}$\\
    $^{1}$School of Computing and Artificial Intelligence, Shanghai University of Finance and Economics\\ 
    $^{2}$Department of Computer Science, University College London, University of London\\
    $^{3}$College of Computer Science and Technology, Guizhou University \\
    {\tt\small yangyitong@stu.sufe.edu.cn, \{wang.yinglin, shuting.he\}@sufe.edu.cn}
}

\let\oldtwocolumn\twocolumn
\renewcommand\twocolumn[1][]{%
    \oldtwocolumn[{#1}{
    \begin{center}
    \vspace{-25pt}
        \centering
        \includegraphics[width=1.0\textwidth]{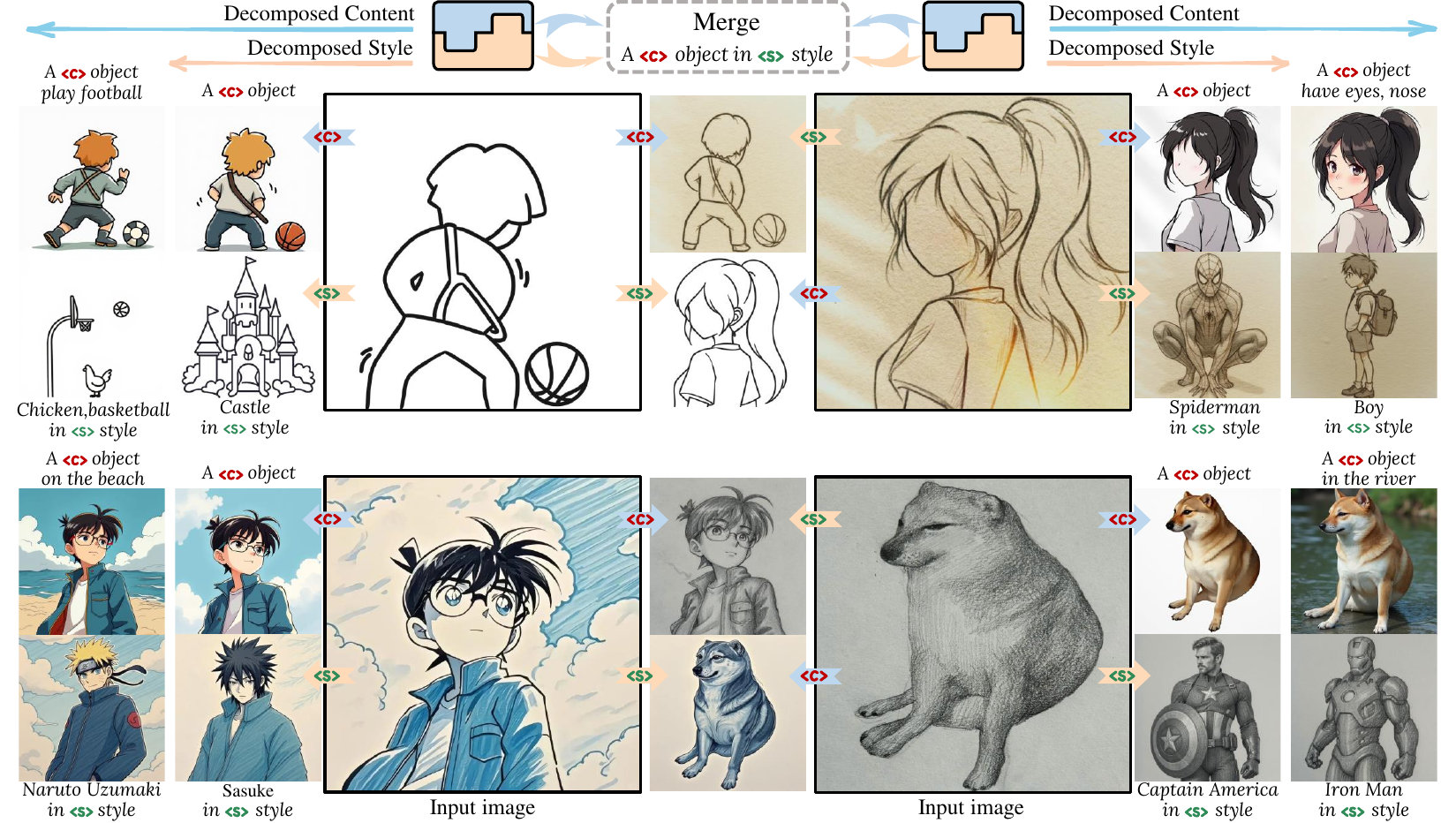}
        \vspace{-20pt}
        \captionof{figure} {SplitFlux is capable of disentangling image content and style, and it can also embed the content into new contexts.}
        \label{fig:teaser}
    \end{center}
    }]
}
\vspace{-10pt}
\begin{document}
\maketitle
\blfootnote{\hspace{-1.8em}$^*$ Corresponding author}

\begin{abstract}
Disentangling image content and style is essential for customized image generation. Existing SDXL-based methods struggle to achieve high-quality results, while the recently proposed Flux model fails to achieve effective content-style separation due to its underexplored characteristics. To address these challenges, we conduct a systematic analysis of Flux and make two key observations: (1) Single Stream Blocks are essential for image generation; and (2) Early single stream blocks mainly control content, whereas later blocks govern style. Based on these insights, we propose \textbf{SplitFlux}, which disentangles content and style by fine-tuning the single stream blocks via LoRA, enabling the disentangled content to be re-embedded into new contexts. It includes two key components: (1) \textbf{Rank-Constrained Adaptation}. To preserve content identity and structure, we compress the rank and amplify the magnitude of updates within specific blocks, preventing content leakage into style blocks. (2) \textbf{Visual-Gated LoRA}. We split the content LoRA into two branches with different ranks, guided by image saliency. The high-rank branch preserves primary subject information, while the low-rank branch encodes residual details, mitigating content overfitting and enabling seamless re-embedding. Extensive experiments demonstrate that SplitFlux consistently outperforms state-of-the-art methods, achieving superior content preservation and stylization quality across diverse scenarios.
\end{abstract}    
\section{Introduction}
\label{sec:intro}
Large-scale text-to-image diffusion models~\cite{flux2024,podell2023sdxl} have recently achieved remarkable improvements in image quality, diversity, and controllability, excelling in tasks such as image editing~\cite{brooks2023instructpix2pix,yang2024prompt,zhang2025context} and style transfer~\cite{wang2024instantstyle,chung2024style,xing2024csgo}. Meanwhile, DreamBooth~\cite{ruiz2023dreambooth} has pioneered a new paradigm for personalized concept generation, inspiring extensive follow-up research~\cite{ye2023ip,voynov2023p+,van2023anti,avrahami2023break,mou2024t2i,sohn2023styledrop} that further advances customized image synthesis. However, these methods often model either the content or the style of an image without effectively disentangling the two, thereby limiting flexible and independent control in generative applications.

Subsequent approaches have further explored this direction based on DreamBooth~\cite{ruiz2023dreambooth} and LoRA~\cite{hu2022lora}, which can be broadly categorized into two groups: content–style disentanglement~\cite{frenkel2024implicit,liu2024unziplora} and content–style integration~\cite{shah2024ziplora,zhong2024multi,roy2025duolora,chen2025consislora,ouyang2025k}. The former identifies specific blocks within diffusion models that can disentangle content and style using a small number of trainable parameters, while the latter requires full fine-tuning of the model or a set of merged parameters to integrate the content and style from different images. However, these approaches rely on the SDXL~\cite{podell2023sdxl} architecture, which struggles to produce high-quality results. Recent methods~\cite{yang2025qr} based on the Flux model~\cite{flux2024} have attempted to integrate different image contents and styles. Although these approaches achieve impressive generative outcomes, several key challenges remain:(1) \textbf{Underexplored Characteristics.} Several studies~\cite{avrahami2025stable,wei2025freeflux} have preliminarily explored the role of different blocks in the Flux; however, their potential characteristics in content–style disentanglement remain largely unexplored. (2) \textbf{Identity Preservation.} The disentangled content frequently exhibits a loss of the subject’s structural and identity features. (3) \textbf{Recontextualization.} The disentangled content is susceptible to overfitting, which hinders its effective re-embedding and flexible application in new contexts.

To address these limitations, we propose \textbf{SplitFlux}, a novel framework that disentangles content and style from a single image, enabling the disentangled content to be re-embedded into new contexts (see Fig.~\ref{fig:teaser}). Inspired by B-LoRA~\cite{frenkel2024implicit}, we perform extensive qualitative and quantitative analyzes of the Flux model to investigate the roles of different layers and derive two key findings: (1) Single Stream Blocks play a crucial role in image generation; (2) Early single stream blocks primarily control image content, whereas later blocks determine image style.

Based on these insights, we fine-tune the Single Stream Blocks of Flux via LoRA, which directly enables content–style disentanglement. However, this process may lead to the degradation of identity and structure in the disentangled content. To mitigate this, we introduce \textbf{Rank-Constrained Adaptation} that compresses the rank while amplifying update magnitude within specific blocks, preventing content leakage into style blocks and preserving identity and structure. Furthermore, to effectively re-embed the disentangled content into new contexts, we propose \textbf{Visual-Gated LoRA}. This module splits the content LoRA into two branches with different ranks and employs a routing function derived from image saliency to ensure that the high-rank branch captures the subject information, while the low-rank branch encodes residual details. The complementary loss encourages the two branches to learn diverse yet complementary representations, enabling the disentangled subject to be seamlessly integrated into new contexts. In summary, our key contributions are as follows:
\begin{itemize}
    \item We identify the functional roles of Flux blocks and derive two key findings that enable effective disentanglement of image content and style.
    \item We propose a Rank-Constrained Adaptation to preserve identity and structure during content–style separation.
    \item We design a Visual-Gated LoRA that embeds disentangled subject information into new contexts.
    \item Extensive experiments demonstrate that our method outperforms existing state-of-the-art approaches in both qualitative and quantitative evaluations.
\end{itemize}
\section{Related Work}
\label{sec:relate_work}

\noindent \textbf{Diffusion Models}.
Diffusion models~\cite{ho2020denoising,song2020denoising,nichol2021improved} have become a cornerstone of generative modeling, achieving impressive results in image generation~\cite{saharia2022photorealistic,ramesh2021zero}, editing~\cite{brooks2023instructpix2pix,sheynin2024emu}, and style transfer~\cite{chen2024artadapter,xing2024csgo}. Early models such as Stable Diffusion 1.5~\cite{rombach2022high} and SDXL~\cite{podell2023sdxl} relied on U-Net~\cite{ronneberger2015u} architectures, which eventually saturated in performance. The Diffusion Transformer (DiT)~\cite{peebles2023scalable} introduced a scalable alternative, exhibiting language-model-like scaling laws~\cite{kaplan2020scaling} and enabling flexible extension to other modalities. Recent DiT-based models, including Flux~\cite{flux2024} and Stable Diffusion 3~\cite{esser2024scaling}, achieve SOTA results in high-resolution generation~\cite{chen2024pixart} and in-context learning~\cite{huang2024group,huang2024context}. Building on the Flux~\cite{flux2024}, we conduct an in-depth study of its intrinsic properties, based on which we propose SplitFlux.
\begin{figure*}[!ht]
    \centering
    \includegraphics[width=\linewidth]{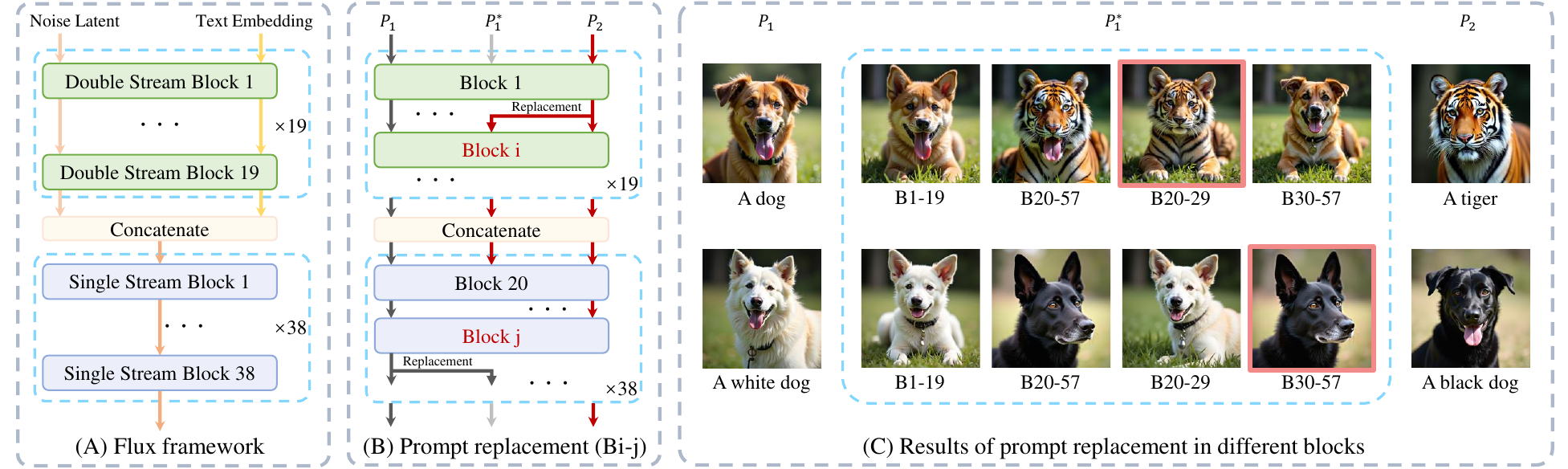}
    \vspace{-20pt}
    \caption{(A) Flux framework. (B) Prompt replacement. $P_1$ and $P_2$ denote distinct prompts, with $P_1^{*}$ being a replication of $P_1$. Replacing $P_1^{*}$ with $P_2$ in a specific block is used to examine the functional contribution of that block. (C) Results of prompt replacement in different blocks. B1–19 denotes injecting $P_2$ into all blocks from Block 1 through Block 19, inclusive.}
    \label{fig:motivation1}
    \vspace{-12pt}
\end{figure*}


\noindent \textbf{LoRA-based Personalized Image Generation}.
Building on diffusion models, customized image generation aims to enable models to produce images aligned with user-specific identities, styles, or concepts. DreamBooth~\cite{ruiz2023dreambooth} pioneered this direction by fine-tuning models with a few subject-specific images to capture identity features, but incurred high computational cost and overfitting risks. To mitigate these issues, Low-Rank Adaptation (LoRA)~\cite{hu2022lora} introduces efficient fine-tuning through low-rank updates to selected layers, greatly reducing memory and computation while preserving fidelity. LoRA-based approaches~\cite{park2024textboost,choi2025memory,gandikota2024concept} have become the dominant paradigm for diffusion model personalization, supporting tasks such as subject-driven synthesis~\cite{wu2024difflora}, style transfer~\cite{liao2023calliffusion,chen2025consislora,li2024diffstyler}, and multi-concept composition~\cite{simsar2025loraclr,zhong2024multi,meral2024clora}. Our task focuses on achieving the disentanglement of image content and style.



\noindent \textbf{Content–Style Disentanglement and Integration}.
Recent parameter-efficient fine-tuning methods have emphasized disentangling and integrating image content and style. Ziplora~\cite{shah2024ziplora} introduced hyperparameters to merge content and style LoRAs for flexible integration, while B-LoRA~\cite{frenkel2024implicit} identified key attention blocks in SDXL to achieve effective disentanglement. Building on these, DuoLoRA~\cite{roy2025duolora} merges LoRAs without retraining via layer-wise priors, and K-Lora~\cite{ouyang2025k} leverages timestep contributions for efficient integration. UnZipLoRA~\cite{liu2024unziplora} further refines disentanglement through prompt and column separation. 
Although effective, these methods often compromise identity and structure. Our approach uses Rank-Constrained Adaptation to preserve both, and Visual-Gated LoRA to embed disentangled content into new contexts.
\section{Preliminary}
\noindent \textbf{Flux}. 
The Flux model is a conditional generative framework based on Flow Matching and the DiT architecture, consisting of 19 double stream blocks and 38 single stream blocks (Fig.~\ref{fig:motivation1} (A)). Input prompts are encoded by CLIP~\cite{radford2021learning} for modulation layers and T5~\cite{raffel2020exploring} for cross-modal interaction. The double stream blocks process text embeddings and image latents independently, while the single stream blocks concatenate image latents with text embeddings to enable joint processing.

\noindent \textbf{LoRA Fine-tuning.} LoRA restricts weight updates to a low-rank subspace. For a linear transformation:
\begin{equation}
\begin{aligned}
h = Wx, \quad  \quad W \in \mathbb{R}^{d_{in}\times d_{out}},
\end{aligned}
\end{equation}
it introduces a low-rank residual update:
\begin{equation}
\begin{aligned}
W^{'}= W+\Delta W, \quad  \quad \Delta W=BA,
\end{aligned}
\end{equation}
where $A \in \mathbb{R}^{r \times d_{in}}$ and $B \in \mathbb{R}^{d_{out} \times r}$, with the rank $r \ll \text{min}(d_{in},d_{out})$. Only $A$ and $B$ are trainable, greatly reducing parameter count. During inference, $\Delta W$ is merged into $W$ without additional cost.

\begin{figure*}[!t]
    \centering
    \includegraphics[width=\linewidth]{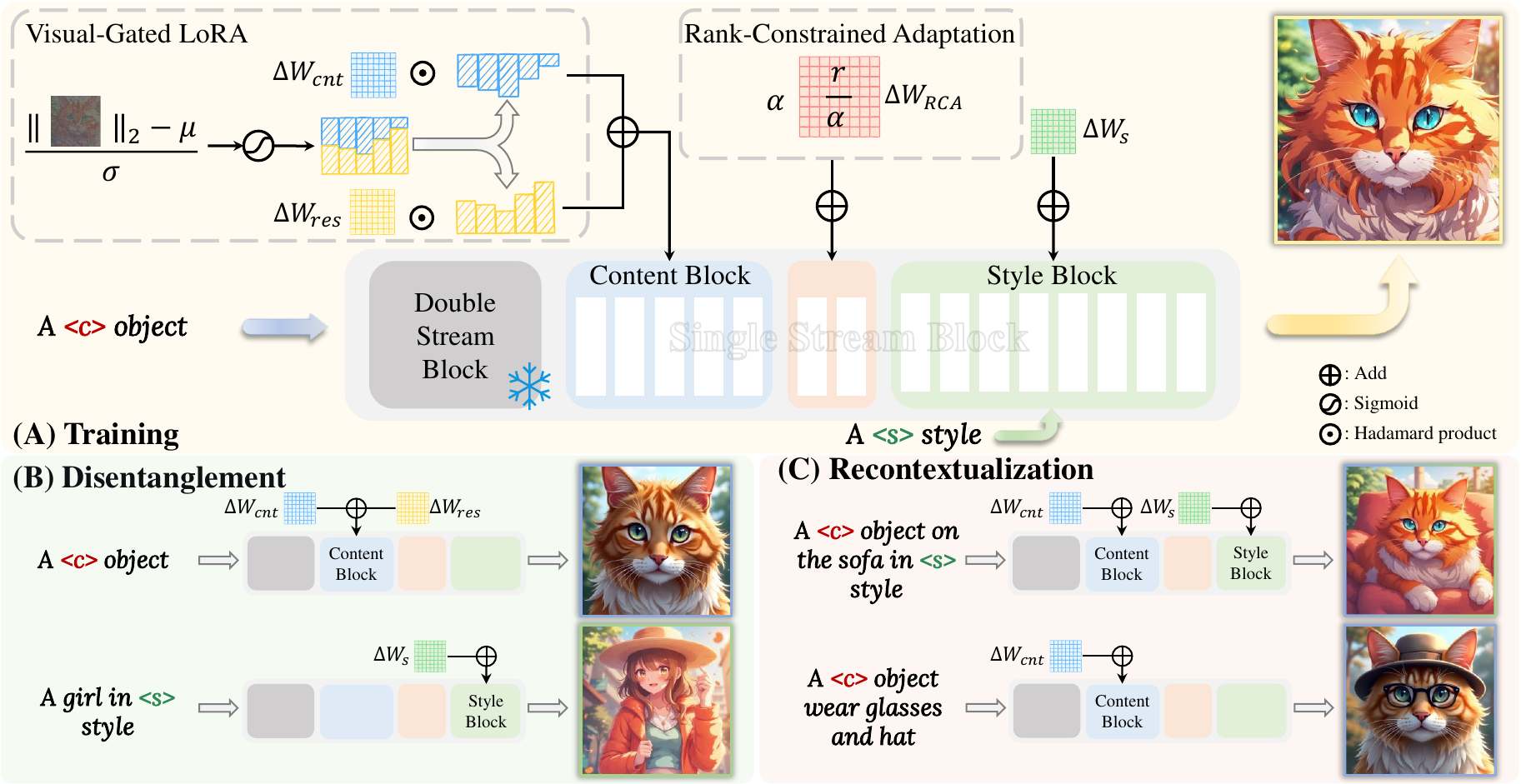}
    \vspace{-20pt}
    \caption{Overview of our method. (A) We train the content and style blocks using two different prompts, enabling the model to disentangle image content and style. We design Rank-Constrained Adaptation (RCA) to preserve content identity and Visual-Gated LoRA (VGRA) to allow the disentangled content to be re-embedded into new contexts. (B) The content and style of an image can be obtained simply by loading the corresponding LoRA. (C) For Recontextualization, we load only $\Delta W_{cnt}$ in the content block.}
    \label{fig:pipeline}
    \vspace{-15pt}
\end{figure*}
\section{Method}
We propose SplitFlux to disentangle content and style from the input image and re-embed content into new contexts (Fig.~\ref{fig:pipeline}). Through extensive analysis of the Flux model, we identify the roles of different blocks in content and style generation (Sec.~\ref{sec:flux}). Based on these insights, we introduce Rank-Constrained Adaptation (RCA) to preserve identity and structure during disentanglement (Sec.~\ref{sec:rca}). Additionally, Visual-Gated LoRA (VGRA) employs saliency-guided rank modulation to ensure that the disentangled content can be seamlessly re-embedded into new contexts (Sec.~\ref{sec:vgra}).
\subsection{Flux Architecture Analysis}
\label{sec:flux}

\begin{table}[t]
\centering
\caption{Ablation study on different blocks. B1–19 denotes injecting $P_2$ into Blocks 1–19 (inclusive); other notations follow the same rule. The percentage represents the probability of successfully injecting the semantics of $P_2$ into the generated results. Content is evaluated using a detection model~\cite{liu2023grounding}, and style is assessed with Qwen3-VL~\cite{Qwen2.5-VL}.}
\small
\setlength{\tabcolsep}{6pt}
\renewcommand{\arraystretch}{0.9}
\resizebox{0.75\linewidth}{!}{
\begin{tabular}{lcc}
\toprule
\textbf{Block replacement} & \textbf{Content set} & \textbf{Style set} \\
\midrule
B1--19  & 0\% & 0\% \\
B20--57 & 100\% & 100\% \\
B20--29 & \cellcolor{cyan!10}\textbf{100\%} & \cellcolor{orange!10}0\% \\
B30--57 & \cellcolor{orange!10}0\% & \cellcolor{cyan!10}\textbf{95\%} \\
\bottomrule
\end{tabular}}
\label{tab:motivation1}
\vspace{-10pt}
\end{table}

While prior works~\cite{chen2024delta,chen2025dit,avrahami2025stable} have examined the roles of different blocks in DiT-based text-to-image models, the functional contributions of individual blocks within the Flux model have not yet been thoroughly explored. Inspired by B-LoRA~\cite{frenkel2024implicit}, we adopt a similar strategy to investigate the critical roles of different blocks in Flux. Unlike SDXL~\cite{podell2023sdxl}, Flux updates its text embeddings dynamically at each block. To accommodate this characteristic, we design two separate branches ($P_1$ and $P_2$) to extract the block-specific text embeddings and inject them into a third branch ($P_{1}^{*}$) to evaluate the influence of each block, as illustrated in Fig.~\ref{fig:motivation1} (B). We use ChatGPT~\cite{chatgpt2023} to generate two separate prompt sets: one for content evaluation and another for style evaluation. Each set contains 10 groups of prompts, and for each group, we apply 10 different random seeds, resulting in a total of 100 images. The content prompt set follows the format: \{\textit{$P_1$: ``a photo of a \{object1\}", $P_2$: ``a photo of a \{object2\}"\}}, while the style prompt set follows the format: \textit{\{$P_1$: ``a photo of a \{color1\} object", $P_2$: ``a photo of a \{color2\} object"\}}. All quantitative results are summarized in Tab.~\ref{tab:motivation1}, with corresponding visualizations shown in Fig.~\ref{fig:motivation1} (C). We draw two key observations:
\begin{enumerate}
    \item \textbf{Single stream blocks are critical for image generation.}  Based on the results in Tab.~\ref{tab:motivation1} and columns 2–3 of Fig.~\ref{fig:motivation1} (C), we observe that injecting semantics into the double stream blocks (Blocks 1–19) yields negligible impact on the final outputs. In contrast, introducing semantics into the single-stream blocks (Blocks 20–57) substantially alters both the content and style of the generated images. These findings highlight the single stream blocks as the key modules responsible for semantically guided image generation.
    \item  \textbf{Early single stream blocks primarily control image content, whereas later blocks determine image style.} Columns 4–5 of Fig.~\ref{fig:motivation1} (C) show that different blocks are responsible for distinct aspects of image generation. Specifically, Blocks 20–29 primarily control the image content, while Blocks 30–57 mainly influence the image style. Injecting semantic information into these blocks directly modifies the corresponding visual attributes of the generated images.
\end{enumerate}

\subsection{Rank-Constrained Adaptation}
\label{sec:rca}
\begin{figure}[!t]
    \centering
    \includegraphics[width=\linewidth]{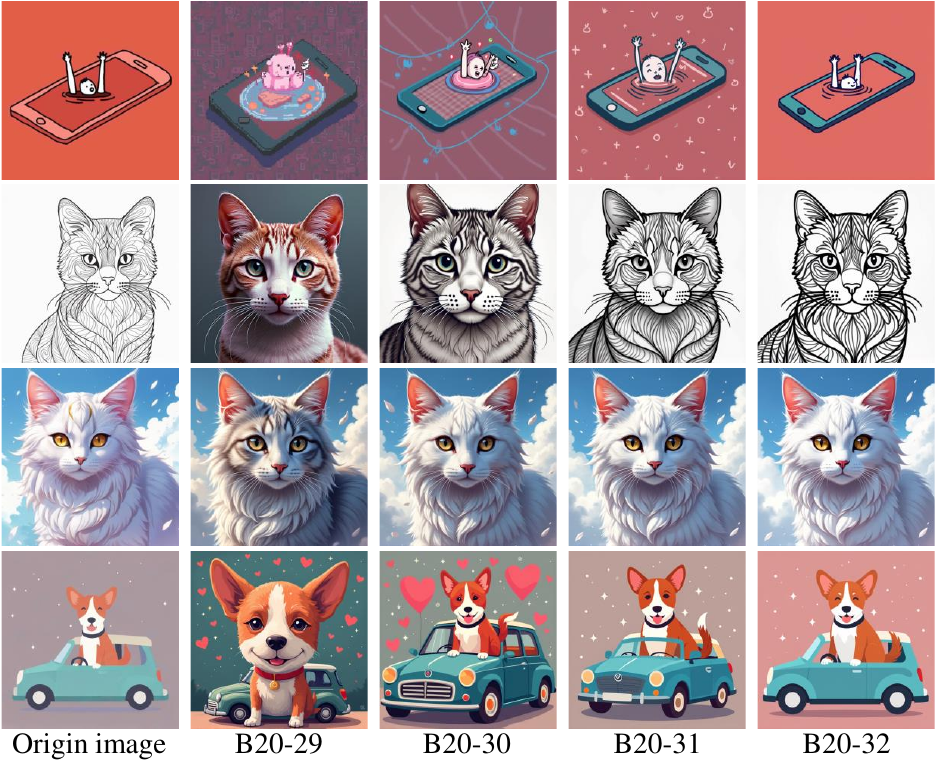}
    \vspace{-20pt}
    \caption{Results of applying block-specific LoRAs for content disentanglement.}
    \label{fig:lora1}
    \vspace{-12pt}
\end{figure}

Based on the above observations, we fine-tune the single stream blocks to obtain the content LoRA ($\Delta W_{c}$) and the style LoRA ($\Delta W_{s}$). By loading the block-specific LoRAs, the model can achieve a certain degree of disentanglement between image content and style, as shown in Fig.~\ref{fig:lora1}. However, when loading the LoRAs corresponding to content blocks, disentangled content often loses its identity or structural integrity, as shown in column 2. Integrating LoRAs from B30, B31, and B32 progressively enhances the preservation of the disentangled content's identity and structure, but it also introduces increasing style features, particularly evident in column 5. This indicates that blocks 30-31 are located at the semantic boundary of the single-stream blocks. At the initial blocks of the style encoding stage, these two blocks simultaneously receive content features from shallow blocks and begin to aggregate high-level style representations. Unconstrained updates to these blocks cause content information to leak into the style subspace, coupling identity-related color and structural cues with stylized features, thereby leading to identity degradation and structural distortion (see the third and fourth columns of Fig.~\ref{fig:lora1}).

To address this issue, we propose Rank Constrained Adaptation (RCA), which jointly constrains the rank and scaling factors of LoRA at the semantic boundary blocks 30-31. This joint constraint restricts the update subspace and amplifies identity-related directions, thereby achieving more stable content–style disentanglement while preserving structural consistency. Specifically, the low-rank update matrix for each layer is given by:
\begin{equation}
\begin{aligned}
\Delta W_{RCA}=\alpha BA, \quad  A \in \mathbb{R}^{\frac{r}{\alpha}\times d_{in}},\quad  B \in \mathbb{R}^{d_{out}\times \frac{r}{\alpha}},
\end{aligned}
\end{equation}
where $\alpha$ is a scaling factor. The forward process can be equivalently expressed as:
\begin{equation}
\begin{aligned}
h=Wx +\alpha\cdot \Delta W_{RCA}x.
\end{aligned}
\end{equation}
By constraining the low-rank subspace of the semantic boundary blocks with 
$\alpha$, content leakage into the style blocks is suppressed. Simultaneously, amplify the update magnitude to compensate for the capacity limitation introduced by the low rank, ensuring that content information is encoded within the content blocks. RCA can be viewed as an implicit regularization method that enables the disentanglement of content and style, thereby reducing identity loss while maintaining the quality of stylized generation.

\noindent \textbf{Prompt Design.} During training, to prevent specific concepts from being bound to unintended tokens, we provide different prompts to different blocks. For the content block, the prompt is ``\verb|A <c> object|", and for the style block, it is ``\verb|A <s> style|". Notably, this fixed prompt format is applied to all images, rather than requiring per-image replacements of ``object" with the subject content and ``\verb|<s>|" with the detailed style, as done in UnZipLoRA~\cite{liu2024unziplora}.

\begin{figure*}[!t]
    \centering
    \includegraphics[width=\linewidth]{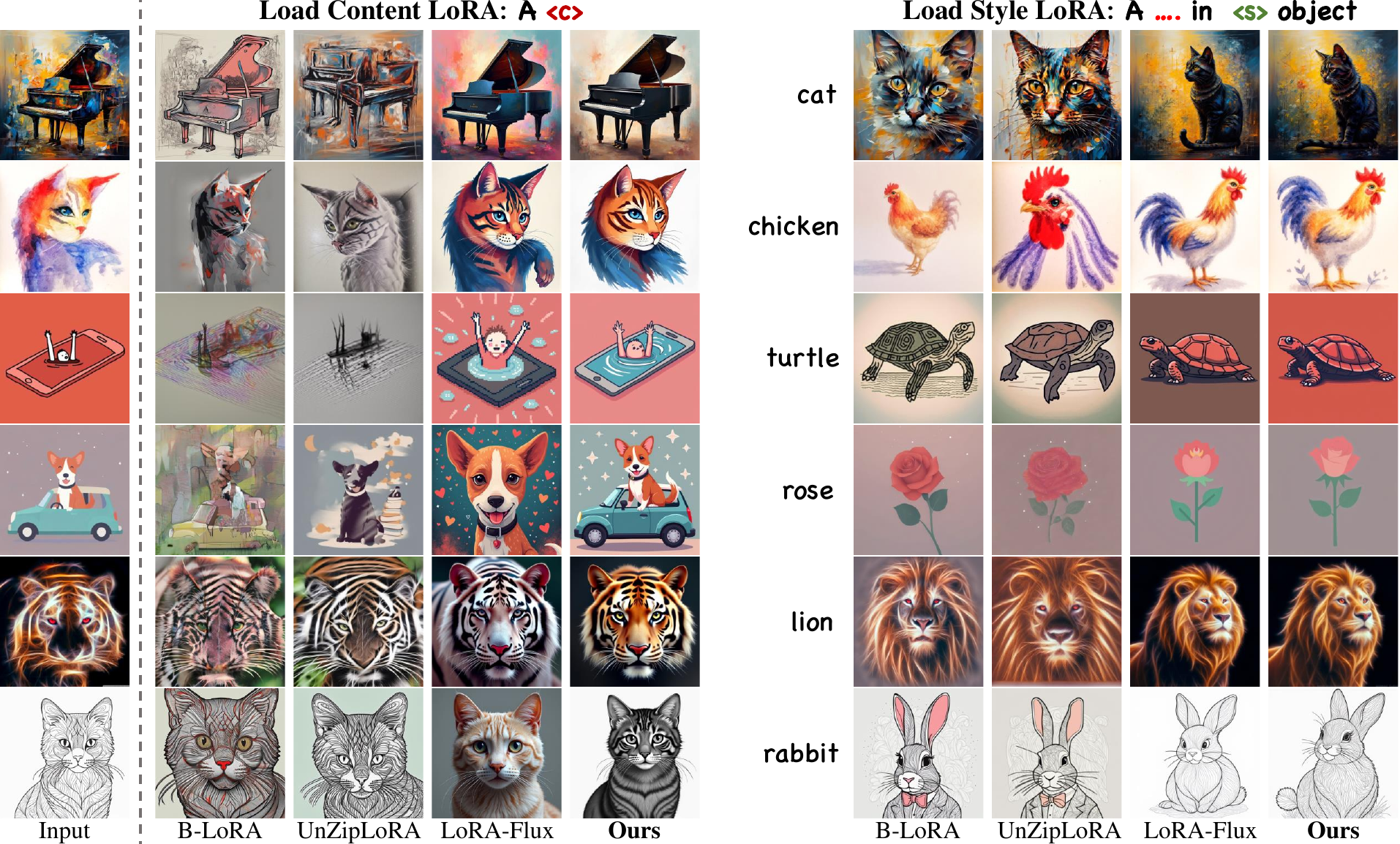}
    \vspace{-20pt}
    \caption{Qualitative comparison for disentanglement. Our results show that our method achieves superior content and style disentanglement, with the disentangled content preserving the original identity and structure better than other methods.}
    \label{fig:e1}
    \vspace{-5pt}
\end{figure*}
\begin{table*}[t]
\centering
\caption{Quantitative comparison of different methods. Disentanglement evaluates the content and style LoRAs separately, while Merger evaluates combinations of content and style LoRAs from different images. \textbf{Bold}: best; \underline{underline}: second best.}
\vspace{-10pt}
\begin{adjustbox}{max width=\textwidth} 
\setlength{\tabcolsep}{5pt}
\renewcommand{\arraystretch}{1.2} 

\begin{tabular}{@{}>{\raggedright\arraybackslash}m{2cm} >{\centering\arraybackslash}m{1.7cm} c c c c c c c c c c c c c c@{}}
\toprule
\multirow{2}{*}{Method} & \multirow{2}{*}{Base Model} & \multicolumn{6}{c}{\textbf{Disentanglement}} 
       & \multicolumn{6}{c}{\textbf{Merger}} 
       & \multicolumn{2}{c}{\textbf{Training}} \\ 
\cmidrule(lr){3-8} \cmidrule(lr){9-14} \cmidrule(l){15-15}
       &            & CLIP-C$\uparrow$ & DINO-C$\uparrow$ & VLM-C$\uparrow$ & CLIP-S$\uparrow$ & DINO-S$\uparrow$ & VLM-S$\uparrow$
       & CLIP-C$\uparrow$ & DINO-C$\uparrow$ & VLM-C$\uparrow$& CLIP-S$\uparrow$ & DINO-S$\uparrow$ 
       & VLM-S$\uparrow$ & Params$\downarrow$ \\ 
\midrule
B-LoRA~\cite{frenkel2024implicit}    & SDXL & 0.760 & 0.547 & 0\% & 0.660 & 0.330 & 6\% & 0.746 & 0.584 & 0\% & \underline{0.695} & 0.368 & 10\% & 56.36M \\
UnZipLoRA~\cite{liu2024unziplora} & SDXL & 0.813 & 0.567 & 15\%& 0.658 & 0.332 & 13.5\%& 0.733 & 0.494 & 8.25\%& \textbf{0.701} & \textbf{0.415} & 19\% & 185.8M \\
LoRA-Flux & Flux & \underline{0.859} & \underline{0.756} & \underline{17.5\%}& \underline{0.665} & \underline{0.358} & \underline{36.5\%}& \underline{0.818} & \underline{0.694} & \underline{14\%}& 0.643 & 0.367 & \underline{23.75\%} & \underline{44.83M} \\
\textbf{Ours} & Flux & \textbf{0.890} & \textbf{0.808} & \textbf{67.5\%} & \textbf{0.666} & \textbf{0.371} & \textbf{44\%}
              & \textbf{0.855} & \textbf{0.765} & \textbf{77.75\%}& 0.657 & \underline{0.368} & \textbf{47.25\%}
              & \textbf{43.65M} \\ 
\bottomrule
\end{tabular}
\end{adjustbox}
\label{tab:comparison}
\vspace{-5pt}
\end{table*}

\subsection{Visual-Gated LoRA}
\label{sec:vgra}
Although RCA effectively decomposes image content and style at the feature level and demonstrates significant advantages in identity-preserving generation tasks, it remains challenging to re-embed the separated content into a new context. This limitation hinders the content representation from achieving flexible transfer and diverse expression in other generative scenarios. To address this limitation, we draw inspiration from Mixture-of-Experts (MoE)~\cite{jacobs1991adaptive} architectures and propose a Visual-Gated LoRA (VGRA) that adaptively routes image tokens to distinct LoRA branches according to their feature saliency. Specifically, given the input representation $x=[E^{T},E^{I}]$, where $E^{T} \in \mathbb{R}^{N \times D}$ and $E^{I} \in \mathbb{R}^{N \times D}$ denote the text and image embeddings respectively, we compute the normalized activation magnitude of each image token as a proxy for visual saliency:
\begin{equation}
s_n=\frac{\|E^{I}_n\|_2-\mu}{\sigma},
\end{equation}
where $E^{I}_{n}$ represents the feature of the $n$-th image token, and $\mu$, $\sigma$ are the mean and standard deviation across all tokens within the image embedding. This normalization ensures that saliency reflects relative activation strength rather than absolute magnitude, emphasizing semantically important regions such as the foreground or the main object. A sigmoid function then transforms the normalized saliency into a differentiable gate:
\begin{equation}
g_i=\text{Sigmoid}(s_i).
\end{equation}

We divide the content LoRA $\Delta W_c$ ($rank=r$) into two distinct LoRA branches, and the updates are computed as their weighted combination:
\begin{equation}
\begin{aligned}
\Delta W_c=g\odot\Delta W_{cnt}+(1-g)\odot\Delta W_{res},
\end{aligned}
\end{equation}
where $\odot$ denotes the Hadamard product. We assign different ranks to the two LoRA branches: the content branch $\Delta W_{cnt}$ ($rank=r^{cnt}$) uses a higher rank to capture the main subject, while the residual branch $\Delta W_{res}$ ($rank=r-r^{cnt}$) uses a lower rank to capture the remaining information. This design is primarily intended to prevent overfitting of image content during training.

\noindent \textbf{Complementary Loss.} To enforce complementarity between $\Delta W_{cnt}$ and $\Delta W_{res}$,  we design a loss that regularizes their parameter-space direction and activation-space position:
\begin{equation}
\begin{aligned}
\mathcal{L}_{\text{comp}} 
&= \underbrace{\big(\|A C^\top\|_F^2 + \|B^\top D\|_F^2\big)}_{\text{direction-level}}
+ \underbrace{\overline{|B A \odot D C|}}_{\text{position-level}}.
\label{eq:decouple_loss}
\end{aligned}
\end{equation}
Here, \(A,B\) and \(C,D\) denote the projection matrices of $\Delta W_{cnt}$ and $\Delta W_{res}$, respectively. This dual constraint ensures that the two LoRA branches learn structurally orthogonal and functionally complementary subspaces. Finally, this loss is incorporated into the default reconstruction loss through the weighting factor $\lambda$.

\section{Experiment}

\subsection{Implementation details}
\noindent \textbf{Datasets.} We collect a total of 40 images for our experiments from B-LoRA~\cite{frenkel2024implicit}, UnZipLoRA~\cite{liu2024unziplora}, and StyleDrop~\cite{sohn2023styledrop}.
\begin{figure*}[!t]
    \centering
    \includegraphics[width=\linewidth]{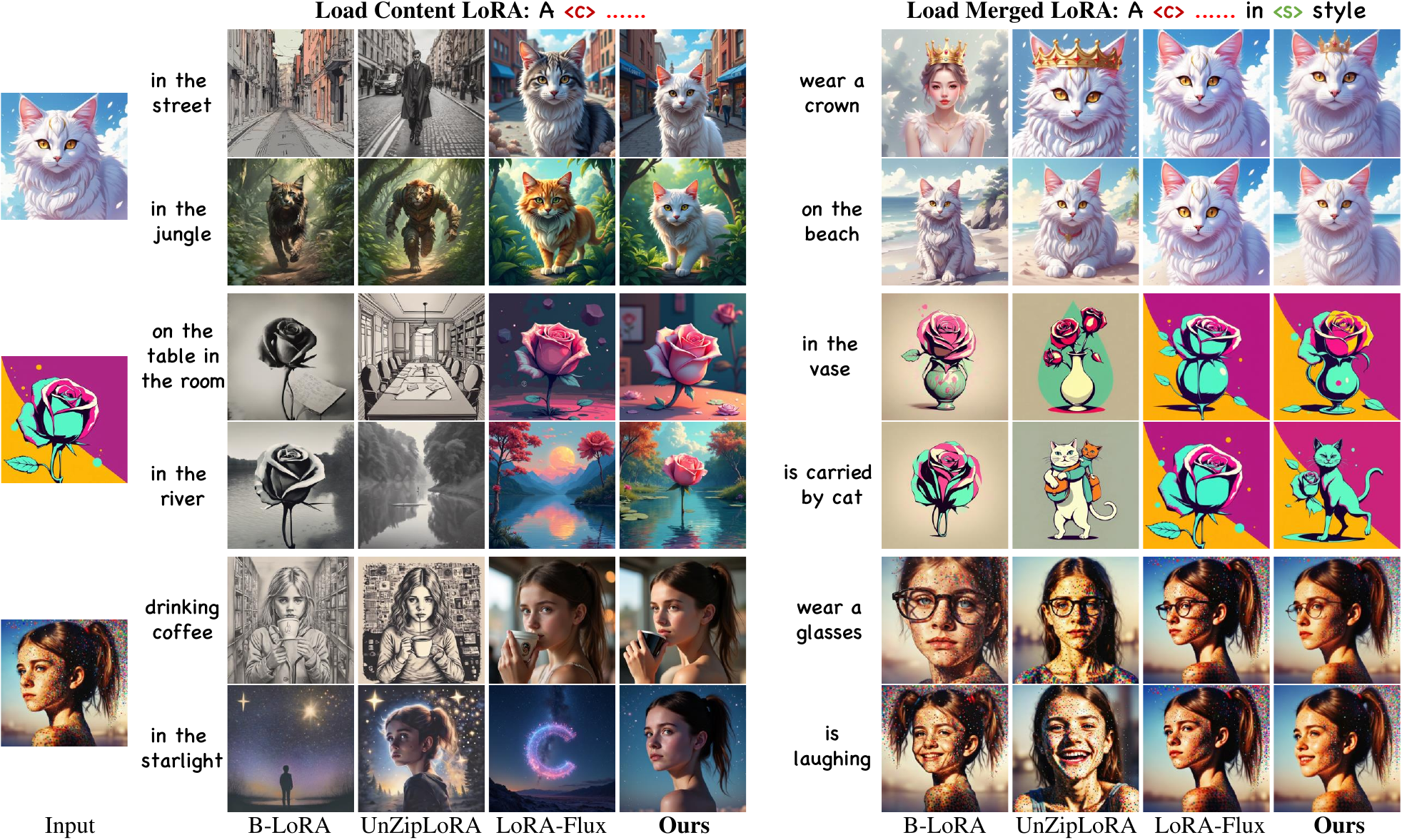}
    \vspace{-20pt}
    \caption{Comparison of different methods for Recontextualization. Our approach flexibly adapts to varying contexts while accurately preserving both the subject and style.}
    \vspace{-12pt}
    \label{fig:e3}
\end{figure*}

\noindent \textbf{Experimental setup.} We use Flux~\cite{flux2024} as the base model for all our experiments. The model is fine-tuned using the Adam~\cite{kingma2014adam} optimizer with a learning rate of 
$1e-4$, a batch size of 1, and trained for 1000 steps. For the content block, the rank of $\Delta W_{cnt}$ ($r^{cnt}$) is set to 48 and that of $\Delta W_{res}$ to 16. For RCA, we set $\alpha=2$ (i.e., rank = 32). For the style block, the rank is set to 64. The weighting factor $\lambda$ in the loss function is set to 0.1. All experiments are conducted on an L20 (48G) GPU.

\noindent \textbf{Compared methods.} We compare our approach with recent methods, UnZipLoRA~\cite{liu2024unziplora} and B-LoRA~\cite{frenkel2024implicit}. We also include LoRA-Flux, which fine-tunes only the single-stream block to provide a comprehensive evaluation. For UnZipLoRA, prompts for training and inference require explicit subject and style descriptions, which is impractical. Therefore, we adopt its main framework prompts as follows:``\verb|A <c> subject in <s> style|"; ``\verb|A <c> subject|"; ``\verb|An image in <s> style|". 

\noindent \textbf{Metrics.} To quantitatively evaluate our method, we compute the cosine similarity between the generated images and their references using DINO~\cite{oquab2023dinov2} and CLIP~\cite{radford2021learning} features, measuring both style (-S) and content (-C) aspects. We also perform a VLM-based (Qwen3-VL~\cite{Qwen2.5-VL}) preference study that mimics a human user study via single-choice questions, assessing content (-C) and style (-S) separately.

\subsection{Comparison Results}
\begin{figure}[!t]
    \centering
    \includegraphics[width=\linewidth]{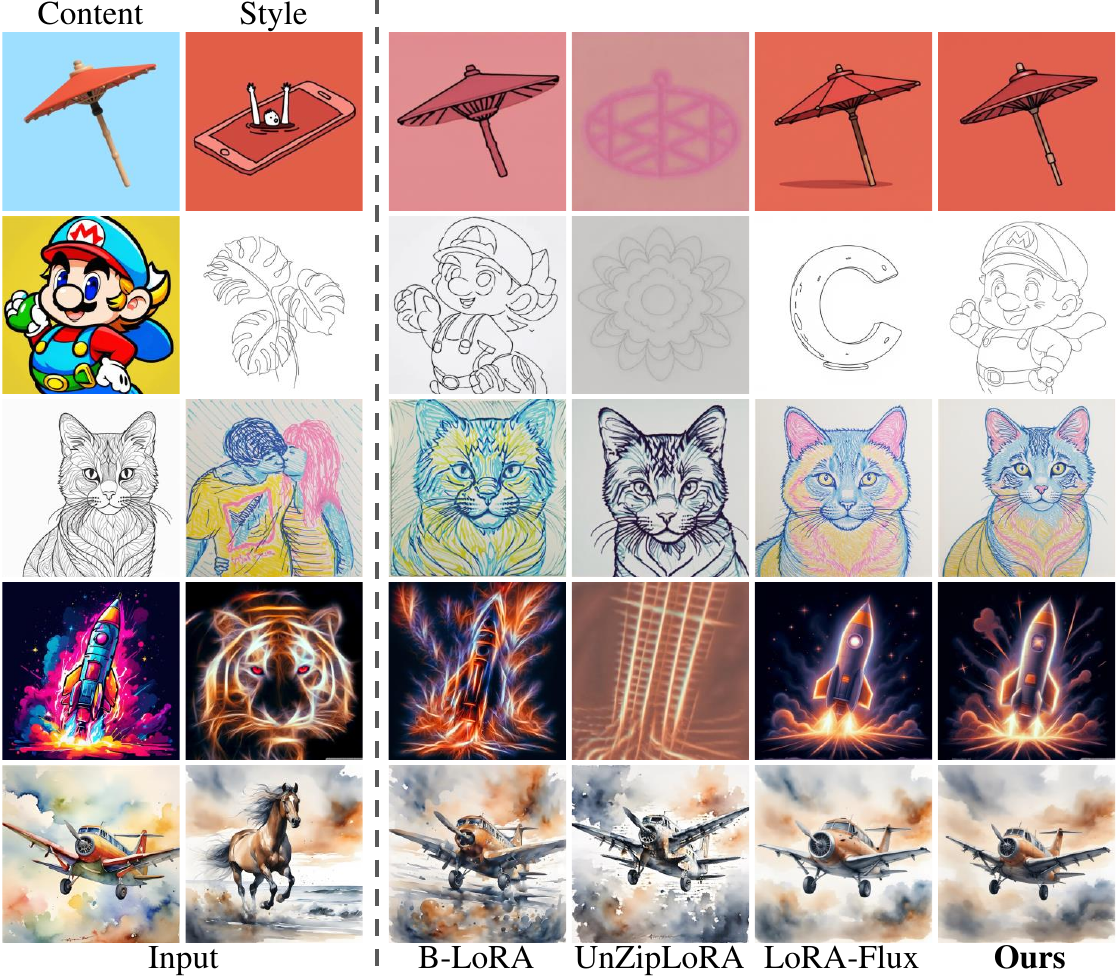}
    \vspace{-20pt}
    \caption{Qualitative comparison for Merger. Compared to other methods, our approach achieves the best combination, maintaining high fidelity in both content and style.}
    \label{fig:e2}
    \vspace{-20pt}
\end{figure}
\noindent \textbf{Quantitative Comparisons}. To comprehensively evaluate our method, we conduct a systematic comparison with existing approaches from two perspectives: Disentanglement (separating content and style LoRAs) and Merger (combining different LoRAs), using multiple metrics (Tab.~\ref{tab:comparison}). For style disentanglement, we generate 10 prompts with ChatGPT and test each style using 5 random seeds. Tab.~\ref{tab:comparison} shows that our method achieves the best performance across multiple metrics, particularly in content preservation, with significant improvements over other methods. For style metrics, our method performs comparably to LoRA-Flux, indicating that the introduction of RCA and VGRA does not compromise style transfer capability. In the Merger task, UnZipLoRA achieves the highest style scores, but at the cost of substantial content loss, as illustrated in Fig.~\ref{fig:e2}. For the VLM-based preference study, our method achieves the best performance in both content and style aspects. Notably, our method uses the fewest training parameters and is far more efficient than UnZipLoRA.

\noindent \textbf{Qualitative Comparisons.} Fig.~\ref{fig:e1} compares disentanglement across methods. B-LoRA partially separates structure but struggles to preserve content, likely due to SDXL’s limitations. UnZipLoRA, extending B-LoRA's block-wise separation to more blocks, further degrades decoupled content. LoRA-Flux achieves effective content disentanglement using the latest Flux model but loses some structural and identity information. Our method better preserves content, achieving the best results (Fig.~\ref{fig:e1}, rows 4–6), and attains style quality comparable to LoRA-Flux while surpassing B-LoRA and UnZipLoRA (Fig.~\ref{fig:e1}, rows 3 and 6). Fig.~\ref{fig:e2} shows that when merging content and style LoRAs, our approach outperforms others in preserving structure, identity, and style.

We further evaluate the ability to transfer content to new contexts (Fig.~\ref{fig:e3}). During transfer, B-LoRA and UnZipLoRA frequently lose the original content, resulting in degraded visual quality. LoRA-Flux may lose color information (rows 1–2) or even the entire content (row 6). When merging LoRAs for context transfer, B-LoRA and UnZipLoRA can perform the transfer but often at the cost of losing the original style (rows 3–4) and identity (rows 5–6). LoRA-Flux lacks a Visual-Gated LoRA, causing content to overfit and hindering effective transfer to new contexts. By contrast, our method flexibly applies content to new contexts while maintaining higher visual quality.

\subsection{Ablation Study}
\begin{table}[t]
\centering
\caption{Quantitative results of the ablation study on RCA.}
\vspace{-6pt}
\renewcommand{\arraystretch}{0.9}
\setlength{\tabcolsep}{4pt}
\small
\begin{tabularx}{\linewidth}{X S[table-format=1.3] S[table-format=1.3] S[table-format=1.3] S[table-format=1.3]}
\toprule
       & {CLIP-C$\uparrow$} & {CLIP-S$\uparrow$} & {DINO-C$\uparrow$} & {DINO-S$\uparrow$} \\ 
\midrule
w/o RCA & 0.859 & 0.665 & 0.756 & 0.358 \\ 
\midrule
\multicolumn{5}{c}{\textbf{Varying $\alpha$; Block 30–31}} \\ 
\midrule
\rowcolor{gray!5}
$\alpha=0$ & 0.842 & 0.645 & 0.709 & 0.337 \\
\rowcolor{gray!5}
$\alpha=4$ & 0.877 & 0.664 & 0.781 & 0.370 \\
\rowcolor{gray!5}
$\alpha=8$ & 0.878 & 0.663 & 0.777 & 0.368 \\
\rowcolor{blue!5}
$\alpha=2$ & \textbf{0.879} & \textbf{0.666} & \textbf{0.784} & \textbf{0.370} \\
\midrule
\multicolumn{5}{c}{\textbf{Varying Block; $\alpha=2$}} \\ 
\midrule
\rowcolor{gray!5}
B30–32 & 0.874 & 0.655 & 0.779 & 0.361 \\
\rowcolor{gray!5}
B30–33 & 0.875 & 0.643 & 0.782 & 0.339 \\
\rowcolor{gray!5}
B30–35 & 0.879 & 0.628 & 0.784 & 0.307 \\
\rowcolor{blue!5}
B30–31 & \textbf{0.879} & \textbf{0.666} & \textbf{0.784} & \textbf{0.370} \\
\bottomrule
\end{tabularx}
\label{tab:ablation}
\end{table}
\begin{figure}[!t]
    \centering
    \vspace{-5pt}
    \includegraphics[width=\linewidth]{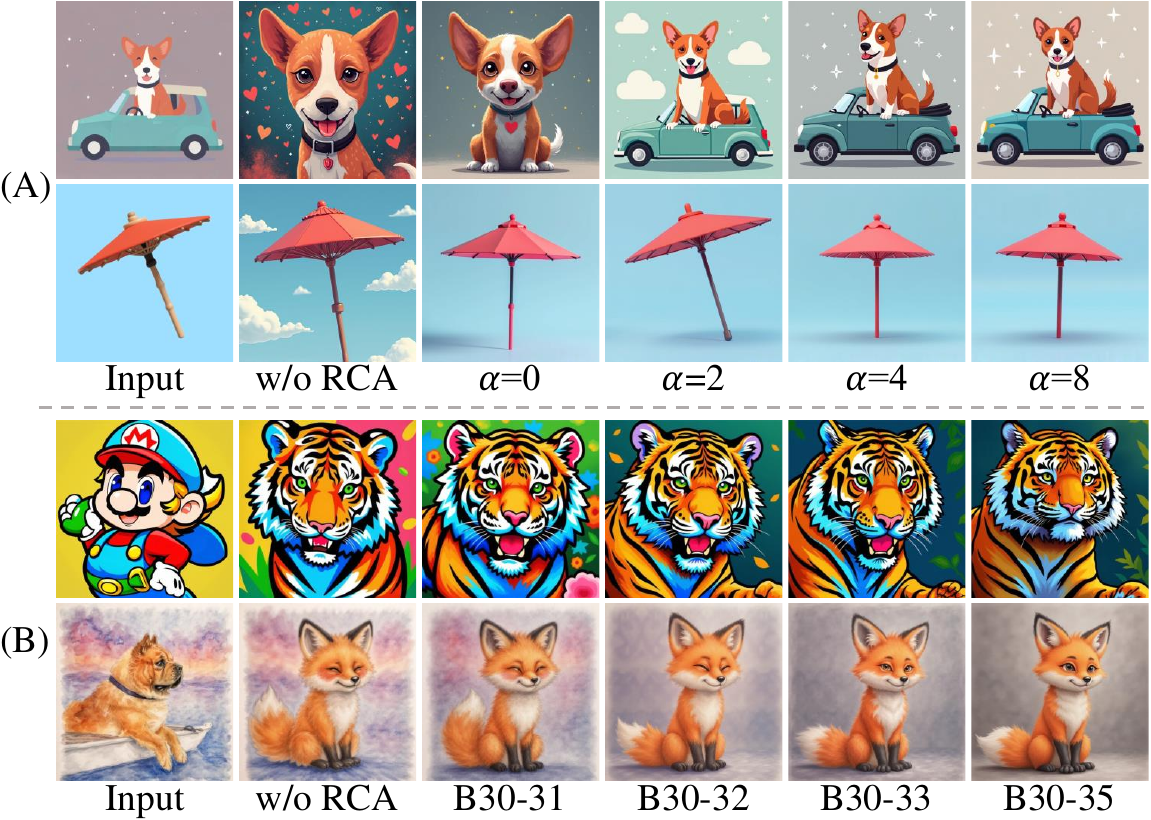}
    \vspace{-20pt}
    \caption{Ablation study on RCA. Our method uses $\alpha=2$ and focuses on Blocks 30–31.}
    \label{fig:a1}
    \vspace{-15pt}
\end{figure}
\begin{figure}[!t]
    \centering
    \includegraphics[width=\linewidth]{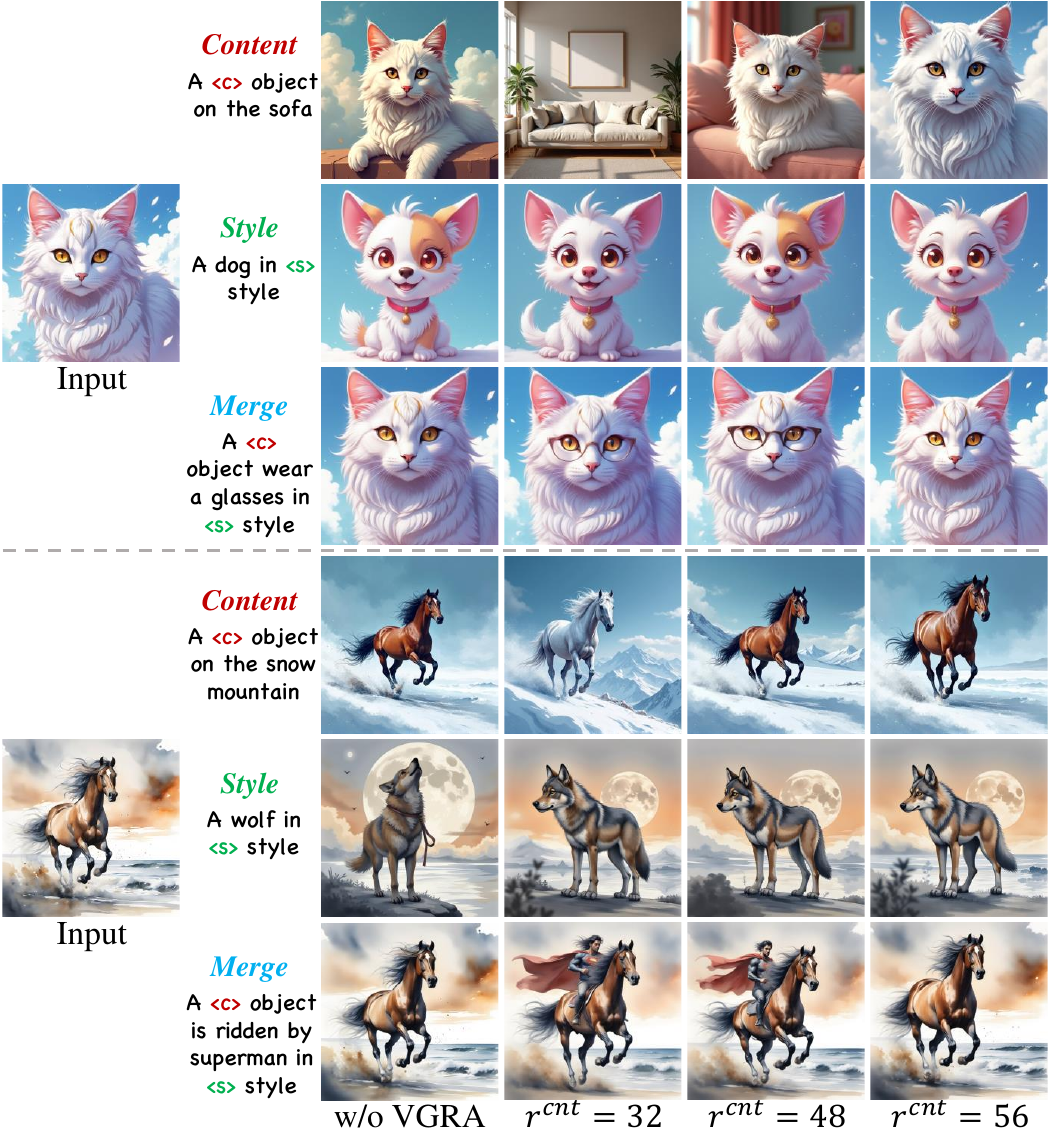}
    \vspace{-20pt}
    \caption{Ablation study on VGRA. Our method uses $r^{cnt}=48$.}
    \label{fig:a2}
    \vspace{-15pt}
\end{figure}
\noindent \textbf{Rank-Constrained Adaptation}. To validate the effectiveness of the proposed RCA method, we conducted both quantitative and qualitative experiments based on the selected block range and $\alpha$, as shown in Fig.~\ref{fig:a1} and Tab.~\ref{tab:ablation}. The results demonstrate that introducing RCA significantly improves content and structural preservation (see the second and fourth columns of Fig.~\ref{fig:a1}~(A)), while having minimal impact on style transfer quality (Tab.~\ref{tab:ablation}). When $\alpha=0$ (i.e., without fine-tuning Blocks 30–31), the information flow between content and style blocks is suppressed, leading to a substantial degradation in both content preservation and style transfer quality (see the third column of Fig.~\ref{fig:a1}~(A) and Tab.~\ref{tab:ablation}). Conversely, when the rank of Blocks 30–31 is too small (i.e., larger $\alpha$), slight content loss can be observed. Furthermore, when the block range selected for RCA becomes excessively large, the quality of style transfer decreases notably, as the reduced number of effective style blocks limits the model’s ability to learn expressive style features (see Fig.~\ref{fig:a1}~(B)).

\noindent \textbf{Visual-Gated LoRA}. Although image content and style can be disentangled without VGRA, the separated content cannot be effectively embedded into new contexts and fails to introduce new semantics to the original image (see Fig.~\ref{fig:a2}, first column). We further conduct an ablation study on the rank of $\Delta W_{\mathrm{cnt}}$ in VGRA. The results show that a larger rank better fits the content information but weakens its adaptability to new contexts (see fourth column), whereas a smaller rank leads to information loss in the disentangled content (see second column). Moreover, as shown in the third and sixth rows of Fig.~\ref{fig:a2}, introducing VGRA does not degrade the quality of style transfer.

\section{Conclusion}
We introduced SplitFlux, which achieves effective content–style disentanglement and enables the re-embedding of disentangled content into new contexts. Through extensive analysis of the Flux model, we identified the important roles of different blocks in content and style generation. Building on these insights, we proposed Rank-Constrained Adaptation to preserve identity during disentanglement and Visual-Gated LoRA to guide content re-embedding via saliency-aware rank modulation. Experimental results demonstrate that our method outperforms the others.
{
    \small
    \bibliographystyle{ieeenat_fullname}
    \bibliography{main}
}


\end{document}